\documentclass[letterpaper]{article}
\usepackage{aaai25} % DO NOT CHANGE THIS
\usepackage{times} % DO NOT CHANGE THIS
\usepackage{helvet} % DO NOT CHANGE THIS
\usepackage{courier} % DO NOT CHANGE THIS
\usepackage[hyphens]{url} % DO NOT CHANGE THIS
\usepackage{graphicx} % DO NOT CHANGE THIS
\usepackage{graphicx}  % DO NOT CHANGE THIS
\usepackage{natbib}  % DO NOT CHANGE THIS
\usepackage{caption}  % DO NOT CHANGE THIS
\usepackage{color}
\usepackage{xcolor}
\usepackage{natbib} % Load natbib before redefining citation commands

\usepackage{float}

\definecolor{mycitecolor}{rgb}{0.21,0.49,0.74}

\makeatletter
\let\oldcitep\citep
\def\citep#1{%
  \begingroup
    \color{mycitecolor}%
    \oldcitep{#1}%
  \endgroup
}
\makeatother

\usepackage{caption} % DO NOT CHANGE THIS AND DO NOT ADD ANY OPTIONS TO IT
\usepackage{graphicx}
\usepackage{amsmath}
\usepackage{amssymb}
\usepackage{amsthm}
\usepackage{booktabs}
\usepackage{algorithmic}
\usepackage[switch]{lineno}
\usepackage{color, colortbl}
\usepackage{multirow}

\usepackage[linesnumbered,ruled,vlined]{algorithm2e}
\usepackage{amsmath, amssymb}

\usepackage{subcaption}
\usepackage{adjustbox}
\usepackage{makecell}

\usepackage{newfloat}
\usepackage{listings}

\def\wrt{\textit{w.r.t.} }
\def\ie{\textit{i.e.}}
\def\eg{\textit{e.g.}}

\usepackage{etoolbox}

\usepackage[capitalize]{cleveref}
\Crefname{section}{Sec.}{Secs.}
\Crefname{table}{Tab.}{Tabs.}
\Crefname{equation}{Eq.}{Eqs.}
\Crefname{figure}{Fig.}{Figs.}
\Crefname{lemma}{Lemma}{Lemmas}
\Crefname{theorem}{Theorem}{Theorems}
\Crefname{definition}{Definition}{Definitions}
\Crefname{hypothesis}{Hypothesis}{Hypothesises}

\DeclareCaptionStyle{ruled}{labelfont=normalfont,labelsep=colon,strut=off} % DO NOT CHANGE THIS
\floatstyle{ruled}
\newfloat{listing}{tb}{lst}{}
\floatname{listing}{Listing}
\title{Improving Generalization of Universal Adversarial Perturbation via \\ Dynamic Maximin Optimization}

\author {
    Yechao Zhang\textsuperscript{\rm 1}, Yingzhe Xu\textsuperscript{\rm 1}, Junyu Shi\textsuperscript{\rm 1}, Leo Yu Zhang\textsuperscript{\rm 2}, \\
Shengshan Hu\textsuperscript{\rm 1}, Minghui Li\textsuperscript{\rm 1}, Yanjun Zhang\textsuperscript{\rm 3}\\
}
\affiliations {
    \textsuperscript{\rm 1}Huazhong University of Science and Technology \\
    \textsuperscript{\rm 2}Griffith University\\
    \textsuperscript{\rm 3}University of Technology Sydney\\
    \{ycz, hynxyz, string1, hushengshan, minghuili\}@hust.edu.cn, leocityu@outlook.com, yanjun.j.zhang@gmail.com
}

\begin{document}

\maketitle

\begin{abstract}

Deep neural networks (DNNs) are susceptible to universal adversarial perturbations (UAPs). These perturbations are meticulously designed to fool the target model universally across all sample classes. Unlike instance-specific adversarial examples (AEs), generating UAPs is more complex because they must be generalized across a wide range of data samples and models. Our research reveals that existing universal attack methods, which optimize UAPs using DNNs with static model parameter snapshots, do not fully leverage the potential of DNNs to generate more effective UAPs.
Rather than optimizing UAPs against static DNN models with a fixed training set, we suggest using dynamic model-data pairs to generate UAPs. 
In particular, we introduce a dynamic maximin optimization strategy, aiming to optimize the UAP across a variety of optimal model-data pairs. We term this approach DM-UAP.
 % Deep neural networks (DNNs) are susceptible to universal adversarial perturbations (UAPs), which are intentionally crafted to deceive them across a wide range of inputs from various classes. Generating UAPs is more complex than creating instance-specific adversarial examples (AEs) due to the need for UAPs to generalize across both samples and models. This paper argue that existing universal attack methods, which optimize UAPs against DNNs with static snapshots of model parameters, do not fully harness the potential of the model to produce stronger UAPs.
% Instead of optimizing UAPs against static DNN models with a fixed training dataset, we suggest a different strategy—dynamic maximin optimization. This method involves optimizing UAPs against a varied and optimal set of model-data pairs, effectively leveraging the DNN's capabilities. We introduce this novel approach as Dynamic Model-Universal Adversarial Perturbation (DM-UAP).
% In each optimization step, we minimize the classification loss by optimizing bounded perturbations to both the data and model parameters. This creates a worst-case scenario for the maximization of UAP, considering the composition of both the input space and parameter space.
DM-UAP utilizes an iterative max-min-min optimization framework that refines the model-data pairs, coupled with a curriculum UAP learning algorithm to examine the combined space of model parameters and data thoroughly. Comprehensive experiments on the ImageNet dataset demonstrate that the proposed DM-UAP markedly enhances both cross-sample universality and cross-model transferability of UAPs.
Using only 500 samples for UAP generation, DM-UAP outperforms the state-of-the-art approach with an average increase in fooling ratio of 12.108\%. 
Our codes are publicly available at \texttt{https://github.com/yechao-zhang/DM-UAP}
% outperforming current state-of-the-art approaches with notable superiority.

%老师，现在篇幅超了，需要精炼，

  % As a result, their generalization capability on unseen samples and models is limited. To tackle this challenge, we propose a novel approach that simultaneously optimizes the data and model used for UAP generation in each mini-batch step. We frame the UAP generation task as a minimax optimization problem. In each optimization step, we minimize the classification loss by introducing bounded perturbations to both the data and model parameters. This creates a worst-case scenario for the maximization of UAP, considering the composition of both the input space and parameter space. To prevent any hard-start issues and expand the search space for these worst-case scenarios, we employ a curriculum strategy that gradually increases the perturbation budgets, which also allows the generalizability of UAP stably increase with the training epochs. Extensive experiments conducted on the widely used ImageNet dataset demonstrate that our method significantly improves the generalization ability of UAP. Furthermore, it outperforms other state-of-the-art methods in terms of both cross-sample universality and cross-model transferability.
\end{abstract}

\vspace{-0.1mm}
\section{Introduction}
Deep Neural Networks (DNNs) represent the cutting edge of computer vision \cite{krizhevsky2012imagenet,simonyan2014very,he2016deep}.
A key strength of DNNs is their ability to identify subtle patterns beyond human perception.
However, this capability of DNNs can unintentionally enable adversarial attacks, where deceptively minor adversarial perturbations of the input---imperceptible to humans---create adversarial examples (AE) that lead to mispredictions of DNNs~\cite{SzegedyZSBEGF13}.

 % Paradoxically, DNNs' capability to identify subtle patterns can unintentionally enable attacks~\cite{SzegedyZSBEGF13,goodfellow2014explaining,madry2017towards}. In these attacks, minute adversarial attacks \cite{SzegedyZSBEGF13}. In such attacks, deceptively minor adversarial perturbations to the input—imperceptible perturbations—unnoticeable to humans—can create be introduced into the input to generate adversarial examples (AEs) that lead (AEs), causing DNNs to the mispredictions of DNNs. make incorrect predictions.

Adding to this concern is the emergence of a notable class of adversarial techniques known as universal adversarial perturbation (UAP), which are designed to induce mispredictions against universal input samples \cite{moosavi2017universal,khrulkov2018art,poursaeed2018generative,shafahi2020universal}.
 % UAPs not only possess the characteristic of \textit{cross-sample universality} but also exhibit another crucial aspect of generalization in adversarial attacks known as \textit{cross-model transferability}. This refers to the ability of (AEs) to retain their effectiveness across various DNN models.
In addition to this characteristic of \textit{cross-sample universality}, UAPs could also possess another element of generalization in adversarial attacks, termed \textit{cross-model transferability}. This denotes the capacity of (AEs) to preserve their efficacy across different DNN models \cite{Hu_2022_CVPR,zhang2024whydoes,10.1145/3581783.3612454}.
% Beyond the \textit{cross-sample universality} characteristic of UAPs, another important generalization aspect of adversarial attacks is \textit{cross-model transferability}, which refers to the ability of adversarial examples (AEs) to maintain their effectiveness across different DNN models.
%Compared to instance-specific perturbations, UAPs pose a greater threat due to their ability to exploit both generalization attributes, generating transferable AEs at scale.
% Together, these capabilities enable UAPs to facilitate the generation of transferable AEs at scale, compromising various DNN architectures and multiple machine learning tasks simultaneously \cite{zhong2022opom,xie2021enabling,ding2021beyond,wang2023improving}.
% As a result, UAPs pose a significant threat, accumulating considerable attention and extensive research.
Together, these capabilities enable UAPs to facilitate the generation of transferable AEs at scale, compromising various DNN architectures and multiple machine learning tasks simultaneously \cite{zhong2022opom,xie2021enabling,ding2021beyond,wang2023improving}. Consequently, UAPs have garnered significant attention and have become a focal point of extensive research in the field.

%, so as to generate transferable AEs at scale. 
%Consequently, UAPs pose a greater threat compared to instance-specific perturbations and have attracted significant attention and undergone extensive research.

% Similarly to DNN model optimization, research formulates the UAP generation task as an optimization problem that maximizes prediction loss on universally perturbed training data and uses stochastic gradient methods with mini-batch training to solve it \cite{shafahi2020universal}.
Recent research aimed at improving the generalization of UAPs has explored different aspects.
 Some focus on analyzing the training data used for UAP generation \cite{zhang2020understanding,li2022learning}, while others reinterpret the gradient aggregation within and between the training batch \cite{shafahi2020universal,liu2023enhancing}. In addition, refinements to training loss have also been explored \cite{zhang2021data}.
% However, these existing works merely optimize UAPs against DNNs with static snapshots of model parameters. %and fail to fully exploit the potential of the underlying model itself to generate more powerful UAPs. 
% However, these existing works fall short by only refining UAPs against fixed DNN parameters, thereby neglecting the dynamic nature of model evolution and failing to leverage the full transformative potential embedded in the underlying models.
However, these existing works fall short by only refining UAPs against fixed DNN parameters, thereby limiting the generalization to static instances of the model without fully harnessing the adaptive capabilities inherent in the model.

% Besides such a standard formulation, a recent work  \cite{li2022learning} takes a step forward by introducing a maximin formulation  which leverages instance-specific adversarial examples  that minimize the loss as the training data for UAPs generation. 
% Such a maximin formulation enforce the resulting UAPs to remain the effectiveness on the antagonistic data, leading to the generation of more dominant UAPs.

% {\color{red} [this is not good, need to summarize them, data perspective, xxx, mini-batch training perspective xxx  ]

% leaving a thorough exploration of dynamic model parameters for generating UAPs an open area of research.
% }

%Our work fills this gap and  contends  that the landscape of model parameters, even within a fixed architecture, can facilitate the generation of robust UAPs.

In this work, we highlight that utilizing the ever-evolving optimized model parameter configurations can enable the generation of stronger and more transferable UAPs.
% In this work, we highlight our finding that  utilizing a diverse set of optimized   configurations of model parameters can enable the generation of stronger and more transferable UAPs. 
% The intuition is that diversity in parameters configurations exposes the generation process to a wider range of model variance, even within a fixed architecture, which fosters  UAPs to capture a broader spectrum of vulnerabilities across models, thereby generalizing to models and data they were not directly trained on.
 The intuition behind this is that exposing the generation process to a wider range of model variance, \textit{even within a fixed architecture}, fosters UAPs to capture a wider spectrum of vulnerabilities between models. This, in turn, allows UAPs to generalize to models on which they were not directly optimized.
% The main objective is to maximize the minimum effectiveness of the UAP across all considered parameter variations.  
% Thus, the generated UAP  is designed to be effective against a range of parameter configurations that a model may adopt, thereby improving its transferability.
Furthermore, we embed the objectives of \textit{cross-sample universality} and \textit{cross-model transferability} into a unified maximin formulation, which involves the composite inner minimization of the model parameters and training data and an outer maximization of the UAP.
%In addition, to further simultaneously improve the  \textit{cross-sample universality} and \textit{cross-model transferability}, we extend this maximin optimization to a more sophisticated one, which involves a composite optimization of both model parameters and training data. 
In other words, we generate UAPs on a wide range of parameter-data pairs that have jointly minimized the classification loss \wrt their respective labels. 
In this way, the generated UAP is optimized to maximize the loss in more antagonistic scenarios from a combined landscape of both parameter and input space. 
% Specifically, we propose a maximin optimization  that  adapts  the model  for each training data point to minimize the classification loss  and utilize the various optimized models  for UAP generation. 
% This adaptive strategy allows us to simulate a diverse set of potential target models that are antagonistic to UAP maximization. 
% \red{in corporated with} the formulation of \cite{li2022learning}, we introduce a dynamic maximin formulation for UAP optimization (DM-UAP), which involves a more sophisticated inner composite minimization over the joint landscape of both model parameters and data, to further improve the generalization of UAP.
% This composite minimization requires the UAP to be effective against a joint landscape of specifically adapted pairs of model and input, thus enhancing its potency and transferability. 
% Nevertheless, such a sophisticated formulation comes with a increasing optimization challenge of navigating the precise composition of each data-model pair and greater convergence difficulty.

To effectively address the optimization challenges associated with this maxmini formulation, we develop an iterative max-min-min optimization framework with dynamic adaptation designs. 
We decouple the inner composite optimization into a two-stage min-min optimization process, where we dynamically optimize a parameter-data pair within each minibatch iteration. 
% To ensure precise navigation of the model-data pair without disrupting the correct distribution learned by the model, we perform model optimization using standard unperturbed data within each iteration, while optimizing the data using the optimized model subsequently.
Additionally, we introduce curriculum UAP learning, in which we first proceed the maximization against ``easy" parameter-data pairs, and gradually increase their optimization spaces along with the training epoch, to ensure a smoother UAP optimization.
In general, the main contributions of our work can be summarized as follows:
\begin{itemize}
    \item We propose a novel maximin formulation, which adapts the model parameters during the optimization process to improve the generalization of UAP. To the best of our knowledge, this is the first exploration of the model parameter landscape for UAP generation.
    
    \item To effectively solve the maximin formulation, we propose a dynamic iterative max-min-min optimization framework, which decouples the composite minimization into a two-stage min-min optimization and leverages curriculum learning to ensure a smooth UAP update.

    \item Extensive experiments on the ImageNet dataset demonstrate the superior generalization of UAPs generated by the proposed DM-UAP compared to the state-of-the-art methods under various attack settings.

\end{itemize}

\section{Related Work}
\subsection{Instance-specific Attacks} 
Instance-specific attacks generate AEs by adding perturbations to individual samples. 
% Existing  instance-specific adversarial attacks can be broadly categorized based on the knowledge of the adversary into three types: white-box attack, query-based attack, and transfer-based  attack.
%\noindent \textbf{White-box attacks.} 
% In a white-box setting,  the adversary possesses complete knowledge of the target model (\ie, its model architecture, and parameters), thus AEs are constructed using gradient directly computed from the target model in either a  single-step \cite{goodfellow2014explaining} or  iterative manner \cite{madry2017towards}. 
In a white-box setting, the adversary has full knowledge of the target model (architecture and parameters), constructing AEs using directly computed gradients, in a single-step \cite{goodfellow2014explaining} or iterative manner \cite{madry2017towards}.
In a black-box setting, the adversary lacks direct access to the target model. 
Alternatively, AEs are crafted against surrogate models, expecting effectiveness against the black-box target model.
Since the discovery of instance-specific AE \cite{szegedy2013intriguing}, research has extensively studied their transferability, exploring loss function refinement \cite{zhao2021success}, input transformations \cite{xie2019improving}, and model ensembles \cite{zhang2024whydoes} to improve the effectiveness. 

\subsection{Universal Adversarial Attacks} 
% Universal attacks aim to craft adversarial perturbations with \textit{cross-sample university}.
% Universal adversarial attacks aim to generate UAPs that can be applied to different instances. 
The seminal work \cite{moosavi2017universal} of UAPs, denoted as UAP, proposed aggregating perturbation vectors obtained from instance-specific attack methods to generate UAPs.
Subsequent work, such as GAP \cite{poursaeed2018generative} and NAG \cite{mopuri2018nag}, introduced the use of generative models for UAP generation.

% Later on, \cite{shafahi2020universal} formulated the UAP generation as an optimization problem that maximizes the averaged prediction loss for the universally perturbed training data, and utilized the stochastic gradient method for generating UAP under the mini-batch training paradigm. 
% Since then, the mini-batch training paradigm has  become a common method  to solve the maximization   and  adopted in  follow-up works~\cite{co2021universal,zhang2020cd,benz2020double,zhang2021data,li2022learning,liu2023enhancing}.

In a later development, SPGD \cite{shafahi2020universal} formulated the generation of UAPs as a maximization of the average prediction loss on universally perturbed training data and used the stochastic gradient method with mini-batch training to solve it. 
Since then, the mini-batch training paradigm has become a common approach to UAP generation in subsequent works \cite{co2021universal,zhang2020cd,benz2020double,liu2023enhancing,10.1145/3474085.3475396}.
Among these, DF-UAP \cite{zhang2020understanding} found that UAPs dominate DNN prediction, and the original samples behave somewhat like random noise after superimposing UAPs.  
This finding inspired AT-UAP \cite{li2022learning}, a more robust UAP achieved by integrating image-specific attacks and universal attacks, to enhance such a dominant effect.
Recently, SGA \cite{liu2023enhancing} 
% examined the gradients accumulation in between different mini-batchs and
proposed to aggregate the noisy gradients obtained from multiple small-batch as a gradient estimation of a large-batch.
However, these methods have not yet sought to improve the generalization of UAPs from the perspective of manipulating the underlying model.
% contended the popular sign operation under the mini-batch training paradigm in UAP generation will cause the  gradient vanishing problem under the small-batch setting and the local optima problem under the large-batch setting. 
% Thus,  they proposed stochastic gradient aggregation (SGA), which utilizes small-batch training to perform multiple inner iterations inside a relatively large batch and aggregates the gradients obtained during these iterations into a one-step gradient estimation.

\section{Methodology}

% In this section, we first introduce the existing problem formulations for UAP optimizations and then we illustrate our novel maximin formulation.
% After that, we present the details of our proposed max-min-min optimization framework of how we solve this maximin formulation.

In this section, we first introduce existing UAP problem formulations and present our novel maximin formulation. After that, we present the details of our proposed max-min-min optimization framework for solving this formulation.

\subsection{Preliminaries}
Suppose $\mathbf{S} $ contains a set of samples $(x, y) \in \mathcal{D}$ drawn from the data distribution $\mathcal{D}$, where $x \in \mathcal{X}$ represents the image feature and $y \in \mathcal{Y}$ denotes its corresponding label. 
The DNN model is depicted as  $f_\theta$, where $f$ is a DNN function and $\theta$ represents the model parameters. 
In general, $\theta$ are optimized through empirical risk minimization (ERM) as follows:
\begin{equation}
\min _\theta \frac{1}{\|\mathbf{S}\|} \sum_{i=1}^{\|\mathbf{S}\|} \mathcal{L} \left(f_\theta \left(x_i\right), y_i\right) \label{eq:training-loss},
\end{equation}
where $\mathcal{L}$ is the loss function (\eg, cross-entropy) for training.

 \textbf{Empirical Risk Maximization.} In a universal adversarial attack, the objective is to craft a single perturbation $\delta$ to fool the well-trained DNN model $f_\theta$ for most images within $\mathcal{X}$. 
In practice, existing work
% \cite{shafahi2020universal,matachana2020robustness,co2021universal,zhang2020understanding,zhang2021data,liu2023enhancing} 
aims to optimize $\delta$ which maximizes the averaged prediction loss for universally perturbed data as follows:

\begin{equation}
\max _{ \|\delta\|_{\infty} \leq \epsilon} \frac{1}{n} \sum_{i=1}^n \mathcal{L} \left(f_\theta \left(x_i+\delta\right), y_i\right) \label{eq:averaged-maximizing},
\end{equation}
where $x_i, ..., x_n \in \mathcal{X}$ are training data used for optimizing the $\delta$, $y_i$ is the corresponding label of the unperturbed input $x_i$ predicted by $y_i=\arg\max f_\theta\left(x_i\right)$, and UAP $\delta$ is a small perturbation constrained by $\ell_\infty$-norm of $\epsilon$.

\iffalse 
Specifically, for generating untargeted UAP,  $\mathcal{J} (\cdot, \cdot)$ is usually set to the same as the training function $\mathcal{L}$, with $y_i =\arg\max f_\theta\left(x_i\right)$.
For generating targeted UAPs, $\mathcal{J} (\cdot, \cdot) =    - \mathcal{L} (\cdot, \cdot)$ with $y_i$ being the chosen target label $y_t$.
\fi

\begin{figure*}[t]
    \centering
    \includegraphics[width=0.9\textwidth]{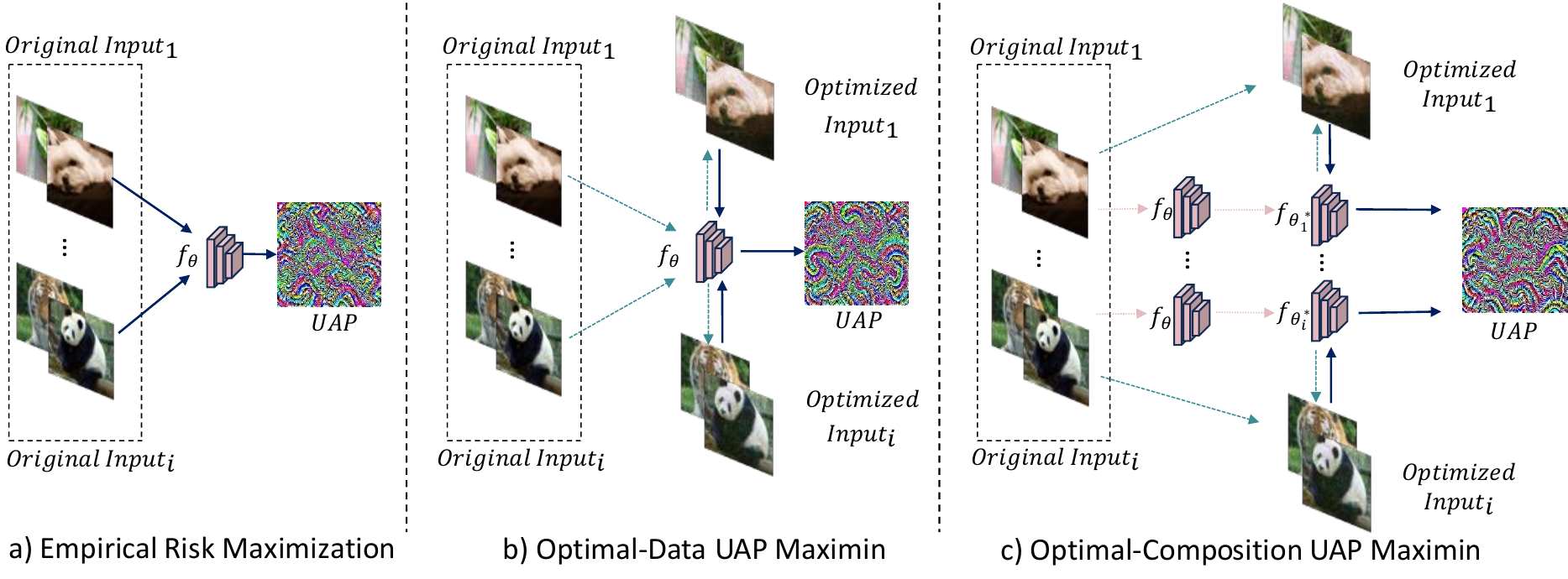}
    \caption{The illustration of different optimization flows: a) \cref{eq:averaged-maximizing} use the original data and model for UAP generation; b) \cref{eq:adversarial-input-maximizing} use the model and original inputs to obtain optimized inputs, then use model and optimized inputs for UAP generation; c)   \cref{eq:dual-maximizing} use original model and original inputs to obtain optimized models and inputs, then use them for UAP generation.}
    \label{fig:flowchart}
\vspace{-3mm}
\end{figure*}

\textbf{Optimal-Data UAP Maximin.} In addition to the standard empirical risk maximization,  \cite{li2022learning} introduces a maximin optimization that leverages image-specific adversarial attacks to produce more effective training inputs. This leads to crafting a UAP based on a tailored training dataset. Formally, this maximin optimization is depicted as follows:

\begin{equation}
\begin{gathered}
\max_{\|\delta\|_{\infty} \leq \epsilon} \frac{1}{n} \sum_{i=1}^n \mathcal{L}(f_\theta(x_i^* + \delta), y_i), \\
\text{s.t. } \forall i, x_i^* = \arg \min_{\|x'-x_i\|_2 \leq r } \mathcal{L}\left(f_\theta(x'), y_i\right).
\label{eq:adversarial-input-maximizing}
\end{gathered}
\end{equation}
This maximin optimization  searches for optimal  data point  \(x_i^*\) in the $r$-bounded $\ell_2$-norm vicinity of each original input \(x_i\), which minimizes the loss for its corresponding label \(y_i\) first. 
Then the UAP \(\delta\) is optimized to  maximize the loss over these optimal data points. 
Compared to \cref{eq:averaged-maximizing}, this formulation constructs antagonistic training data for maximization, compelling the UAP to avoid trivial solutions that may not effectively deceive the model \(f_\theta\).
% Note that the inputs to \(f_\theta\) for the  minimization task (data optimization) and the  maximization task (UAP optimization) are different. This difference ensures that the UAP is crafted against the most adversarial instances of the data within the specified perturbation budget, thereby enhancing the effectiveness of the UAP.

% Generally, for generating the untargeted UAP,  $\ell (\cdot, \cdot)$ is set to be the exact traing loss function of $\theta$, \ie, $\ell$ and $y$ is the corresponding label of the unperturbed input $x \in \mathbb{R}^d$,  predicted by $y=\arg\max f_\theta\left(x\right)$. 

% For targeted attack, $\mathcal{L} (\cdot)$  denotes the designed loss used for generating adversarial perturbation $\delta \in \mathbb{R}^d$, which is restricted by $\ell _p$ norm with perturbation budget $\epsilon$ to ensure the adversarial example $x+\delta$ to be indistinguishable from $x$. 
% In particular, $\delta$ are usually generated from gradient information from the well-trained target CNN model  (\ie, white-box scenario) or surrogate model (\ie black-box scenario).

\subsection{Problem Formulation}

In contrast to previous formulations (\cref{eq:averaged-maximizing,eq:adversarial-input-maximizing}) that optimize UAP against a static model \( f_\theta \)—whether for a direct white-box attack or through a surrogate model in transfer-based attacks—we propose optimizing across multiple variants of the model parameter \( \theta \) to enhance the potency of the UAP.

\textbf{Optimal-Parameters UAP Maximin.}
Inspired by the optimal-data UAP  maximin concept in \cref{eq:adversarial-input-maximizing}, we introduce a parallel optimization as follows:

\begin{equation}
\begin{gathered}
\max_{\|\delta\|_{\infty} \leq \epsilon} \frac{1}{n} \sum_{i=1}^n \mathcal{L}(f_{\theta_i^*} (x_i + \delta), y_i), \\
\text{s.t. } \forall i, \theta_i^* = \arg \min_{\|\theta'-\theta \|_2 \leq \rho}   \mathcal{L}\left(f_{\theta'} \left(x_i\right), y_i\right). \label{eq:adversarial-parameter-maximizing}
\end{gathered}
\end{equation}
% Unlike \cref{eq:adversarial-input-maximizing}, this  optimization finds the optimal model parameters \( \theta_i^* \)  that minimize the loss for each data point \( (x_i,  y_i )\) inside the \(\rho\)-bounded \(\ell_2\)-norm vicinity of original  parameters \( \theta \). 
% Then, the UAP \( \delta \) is optimized to  maximize the average loss across all perturbed data points \( (x_i + \delta, y_i)\), where each data point is evaluated using its respective model variant \( f_{\theta_i^*} \). 
% This formulation ensures that  UAP is effective in increasing loss across different  variations of the model, thus fostering a more robust adversarial effect anticipated to generalize better across various models.
Unlike \cref{eq:adversarial-input-maximizing}, this optimization finds the optimal model parameters \( \theta_i^* \) that minimize the loss for each data point \( (x_i, y_i) \) within a \( \rho \)-bounded \( \ell_2 \)-norm vicinity of the original parameters \( \theta \). The UAP \( \delta \) is then optimized to maximize the average loss across all perturbed data points \( (x_i + \delta, y_i) \), evaluated using their respective model variants \( f_{\theta_i^*} \). This ensures UAP effectively increases loss across different model variations, promoting a robust adversarial effect expected to generalize better across models.

% it necessitates the discovery of perturbations with broad efficacy, circumventing trivial but non-generalizable solutions. 
% Moreover, the maximization takes into account the average loss across diverse model conditions, and data instances.

\textbf{Optimal-Composition UAP Maximin.} 
Building on the optimal-parameters UAP maximin, we propose a more advanced formulation that includes a composite inner minimization over both model parameters and data points:
\begin{equation}
\begin{gathered}
\max_{\|\delta\|_\infty \leq \epsilon} \frac{1}{n} \sum_{i=1}^n \mathcal{L}(f_{\theta_i^*} (x_i^* + \delta), y_i),
\\\text{s.t. } \forall i, (\theta_i^*, x_i^*) = \arg \min_{\substack{\|\theta' - \theta\|_2 \leq \rho \\ \|x' - x_i\|_2 \leq r}} \mathcal{L}(f_{\theta'}(x'), y_i).\label{eq:dual-maximizing}
\end{gathered}
\end{equation}
In this context, each \( ( \theta_i^*, x_i^* ) \) pair is simultaneously optimized to reduce the loss for \( y_i \), while remaining within their respective neighborhoods defined by \( \rho \) and \( r \). This sets up a more challenging condition for the UAP \( \delta \), which requires it to maximize loss against the best-modified models and their corresponding inputs. Both \cref{eq:adversarial-input-maximizing,eq:adversarial-parameter-maximizing} represent special cases of this formulation. \cref{fig:flowchart} illustrate the optimization flow of \cref{eq:averaged-maximizing,eq:adversarial-input-maximizing,eq:dual-maximizing}.

% In this sophisticated max-min framework, each \( \theta_i^*,  x_i^* \) pair are jointly optimized to minimize the loss for \( y_i \), again within respective prescribed neighborhood defined by \(\rho\) and \(r \). 

% This composite minimization creates a more antagonistic scenario, as the UAP \( \delta \) must  maximize the loss against a combined landscape of specially adapted models and their optimal inputs. 
% The formulation is designed to produce a UAP capable of inducing misclassifications under a wider range of adversarial conditions, potentially increasing its potency against unseen models and instances.
% potentially increasing its potency against unseen models and instances.
% The formulation is designed to produce a UAP capable of inducing misclassifications under a wider range of adversarial conditions.

\subsection{Dynamic Maximin UAP:  An Iterative Max-Min-Min Optimization Framework}
Tackling \cref{eq:dual-maximizing} is a non-trivial task. First, the minimization process involves a dual-layered optimization that requires accurate navigation through two high-dimensional spaces. Second, the maximization phase must be flexible enough to cope with a wide range of combined landscapes concerning both data and the model parameters, which poses greater optimization challenges compared to \cref{eq:averaged-maximizing,eq:adversarial-input-maximizing,eq:adversarial-parameter-maximizing}. To overcome these difficulties, we introduce an iterative max-min-min optimization framework. This framework separates the composite minimization into a two-stage min-min optimization and employs curriculum learning for a smoother UAP maximization, as detailed below.

% As the inner minimization process adjusts the model parameters and data,  the maximization step must simultaneously account for these adaptations and steer the optimization towards a perturbation that maintains its effectiveness. 
% This is akin to hitting a moving target – as soon as an effective $\delta$ is found, the model parameters or data may change, rendering the previously optimal $\delta$ less effective.

\noindent\textbf{Dynamic Strategy for mini-batch Training.} 
Intuitively, \cref{eq:dual-maximizing} suggests a straightforward implementation: find all optimal pairs $(x_i^*, f_{\theta_i^*})$, use them as a fixed training set to update UAP through mini-batch training with stochastic gradient ascent. 
However, this is highly inefficient due to the significant storage and memory required for forward-backward propagation across multiple models $f_{\theta_i^*}$.  To address this, we adopt a dynamic strategy: during each mini-batch iteration, we optimize a mini-batch of images with a shared model on the fly, \ie, each random mini-batch of images $X_B$ with the initial model $f_\theta$ induces a composition $(X_B^*, f_{\theta^*})$. After that, the perturbed mini-batch data $X^*_B + \delta$ are propagated forward and backward in parallel through $f_{\theta^*}$ to obtain the update gradient for $\delta$.

% 这个标题可能要放在下一段
\noindent\textbf{Sequential Model and Data Optimization.}  
As discussed earlier, within each mini-batch iteration, we decouple the composite minimization of the data and model parameters into sequential min-min optimization steps. Specifically, 
\begin{itemize}
    \item We first perform model optimization on the mini-batch images $X_B$ to obtain the optimized model $f_{\theta^*}$. 
    \item Then, we use the optimized model $\theta^*$ to perform data optimization to obtain $X_B^*$.
\end{itemize}
The rationale for this ``optimize model first, data second'' approach is to ensure that the model is always updated using unperturbed data. This is because adversarially constructed images $X_B^*$ may not belong to the correct feature distribution $\mathcal{X}$. Overfitting the model to adversarially constructed images $X_B^*$ could result in significant underfitting of the original distribution \cite{dong2020adversarial, xu2022adversarially}. 

\noindent\textbf{Model Optimization.} To ensure the effectiveness of resulting UAP against the original model $f_\theta$, the optimized model parameters $\theta^*$ should not deviate much from the origin $\theta$. However, standard gradient optimizers cannot guarantee precise adherence to the $\ell_2$-norm constraint. To mitigate this, we leverage the normalized gradient of parameters when updating  $\theta^*$. Concretely, we update $\theta^*$  as follows:

\begin{equation}
\theta^*  \leftarrow \theta^* - \alpha_m \frac{\nabla_\theta \frac{1}{B}  \mathcal{L}(f_{\theta^*}(X_B),Y_B)} {\|\nabla_\theta \frac{1}{B}  \mathcal{L}(f_{\theta^*}(X_B),Y_B)\|_2}, \label{eq:theta-update}
\end{equation}
where $Y_B$ are the corresponding labels of  images $X_B$, $\alpha_m$ is the step size of model optimization.  
% Note that we do not apply a projection step on the model parameters, as commonly done in data optimization, because projecting the model parameters may potentially damage important parameters like Batch Normalization. % 老师，我回复了
% Instead, we optimize $\theta^*$ from the origin $\theta$. 
Given the adversarial budget $\rho_t$ at the $t$-th epoch, we update $\theta^*$ for $K_m$ steps and set the step size $\alpha_m$ as $\frac{\rho_t}{K_m}$ to ensure that $\theta^*$ remain within the $\rho_t$-bounded neighborhood of $\theta$.

\noindent\textbf{Data Optimization.}
As discussed earlier, the data optimization is conducted in condition to the optimized model $f_{\theta^*}$ inside each mini-batch iteration. 
Specifically, we update the data using the standard $\ell_2$-norm targeted  PGD  as follows:
\begin{equation}
x^* \leftarrow \Pi_{r_t}(x^* - \alpha_d \cdot \frac{\nabla_x \mathcal{L}(f_{\theta^*}(x^*), y)}  {\|\nabla_x \mathcal{L}(f_{\theta^*}(x^*), y) \|_2}), 
 \label{eq:x-update}
 \end{equation}
where $x^*$ denotes a sample in mini-batch $X_B^*$,   and $\Pi(\cdot)$ is the projection function that ensures the data perturbation does not exceed the budget of $r_t$. 
We set the step size of data optimization $\alpha_d$  as $1.25 \times \frac{r_t}{K_d}$. 
\cref{eq:x-update} is executed for all the samples in $X_B^*$ in parallel, and  repeated for $K_d$ steps.

\begin{algorithm}[t!]
\SetAlgoLined
\KwIn{Data $X = {x_1, \ldots, x_n}$, model $f$ with parameters $\theta$, number of epochs $T$, mini-batch size $B$.}
\KwIn{UAP maximum perturbation magnitude $\epsilon$, initial learning rate $\gamma$.}
\KwIn{Model maximum neighborhood size $\rho$, number of model optimization steps $K_m$.}
\KwIn{Data maximum neighborhood size $r$, number of data optimization steps $K_d$.}
\KwOut{Universal Adversarial Perturbation (UAP) $\delta$}

Initialize $\delta \sim \mathcal{U}(-\epsilon, \epsilon)$;

\For{$t = 1$ to $T$}{
    Compute the perturbation magnitude $\rho_t = t \times \frac{\rho}{T}$ and step size $\alpha_m = \frac{\rho_t}{K_m}$ for model optimization in epoch $t$;
    
Compute the perturbation magnitude $r_t = t \times \frac{r}{T}$ and step size $\alpha_d = 1.25 \times \frac{r_t}{K_d}$ for data optimization in epoch $t$;

\For{mini-batch $X_B \in X$}{

    $Y_B = \arg\max f_\theta(X_B )$\;

    Initialize $\theta^* \leftarrow \theta$\;
    Initialize $X^*_B \leftarrow X_B$  \;

    \For{$k = 1$ to $K_m$}{
        % Compute $g_m = \frac{1}{B} \nabla_\theta \mathcal{L}(f_{\theta^*}(X_B^*), Y_B)$ \;
        
        Update model parameters  with \cref{eq:theta-update}\;
    }

    \For{$k = 1$ to $K_d$}{
        % Compute $g_d = \nabla_x \mathcal{L}(f_{\theta^*}(X_B^*), Y_B)$\;
        
        Update data with \cref{eq:x-update} (\textit{in parallel})\;
    }

    Update UAP with \cref{eq:adam} \;
}
}
\Return{$\delta$};
\caption{Dynamic Maximin UAP (DM-UAP) with Curriculum Learning}
\label{algo:CM-UAP}
\vspace{-1mm}
\end{algorithm}

\noindent\textbf{Curriculum  UAP Learning.} 
The outer maximization of updating UAP is challenging as it requires accumulating gradients from various parameters and inputs, all of which are constantly changing in high-dimensional neighborhoods. To address this, we leverage the concept of curriculum learning \cite{bengio2009curriculum}.
% For curriculum UAP learning, we adopt a progressive approach by initially training the inner min-min optimization of $\theta^*$ and $X_B^*$ in smaller neighborhoods during the early training epochs. Then, we gradually increase the neighborhood size to encompass a larger region.
% Specifically, over a total of $T$ epochs, we set the neighborhood sizes in the $t$-th epoch as $\rho_t = t \times \frac{\rho}{T}$ and $r_t = t \times \frac{r}{T}$ for model optimization and data optimization, respectively. This allows the UAP to progressively strengthen its effectiveness with better convergence.
% By employing the curriculum learning strategy, we facilitate a smoother UAP optimization process, enabling it to  effectively adapt to the dynamic composition of  the model and data.
We adopt a progressive approach by initiating inner-min-min optimization of $\theta^*$ and $X_B^*$ in smaller neighborhoods during the early training epochs. Then, we gradually increase the neighborhood sizes to encompass larger regions over the course of the training.
% Concretely,  we approach the inner min-min optimization of $\theta^*$ and $X_B^*$ in smaller neighborhoods during the early epochs. Then, we gradually increase the neighborhood size to encompass larger regions.
In each epoch $t$ out of a total of $T$ epochs, we set the neighborhood sizes as $\rho_t = t \times \frac{\rho}{T}$ and $r_t = t \times \frac{r}{T}$, respectively. 
% We believe this progressive approach strengthens the UAP's effectiveness improves convergence, and facilitates a smoother UAP optimization process that effectively adapts to the dynamic composition of the model and data.

% To handle the highly dynamic inner minimization in our formulation (\cref{eq:dual-maximizing}),  instead of using the SGD incorporated with sign function as in previous works \cite{liu2023enhancing}, we employ the Adam as our UAP optimizer. 
% Adam adjusts learning rates and momentum to adapt to changing optimization landscapes, and  is suitable for non-stationary objectives, which allows it to update UAP in a way that is adaptive to  the evolving dynamics of the model parameters and data. 
% Thus, the  UAP update in each iteration is formulated as follows:

To better update UAP in a dynamic environment, instead of using SGD with the sign function as in previous works \cite{liu2023enhancing}, we use Adam as the optimizer. Since Adam adjusts learning rates and momentum to adapt to the optimization landscapes, which makes it suitable for such non-stationary objectives, the UAPs are updated adaptively to the evolving dynamics of the model and data. Formally, the UAP  is updated in each iteration as follows:

\begin{equation}
\begin{gathered}
\delta = Adam(\nabla_\delta \mathcal{L}(f_{\theta^*}(X_B^* + \delta), Y_B), \gamma) \\
\delta = \min(\max(\delta, -\epsilon), \epsilon) \label{eq:adam}
\end{gathered}
\end{equation}
where $\gamma$ is the initial learning rate, and the min-max operation  is conducted to ensure the valid $\ell_\infty$-norm constraint on the generated UAP. 
The entire framework of our proposed DM-UAP method is summarized in \cref{algo:CM-UAP}.

% \begin{table*}[!ht]
%     \centering
%     \caption{Target Attack,10000 photos,torward category 805 of ImageNet}
%     \scalebox{0.870}{
%     \begin{tabular}{|l|l|lllll|}
%     \hline
%         method & Number of  Samples & Alexnet & Googlenet & VGG16 & VGG19  & ResNet152  \\ \hline
%         GAP & 10000 & 50.70 & 70.00 & 73.68 & 63.37 & 66.87 \\ 
%         SPGD & 10000 & ~ & ~ & ~ & ~ & ~ \\ 
%         SGA & 10000 & \textbf{62.95} & 76.13 & 86.48 & 83.17 & 83.17 \\ 
%         DM-UAP(Ours) & 10000 & 59.92 & \textbf{77.82} & \textbf{89.47} & \textbf{86.14} & \textbf{85.61} \\ \hline
%     \end{tabular}
%     }

%     \label{white-box-targeted}
% \end{table*}

\section{Experiment}

\noindent\textbf{Setup:} Following \cite{moosavi2017universal,liu2023enhancing}, we randomly select 10 images from each category in the ImageNet training set, resulting in a total of 10,000 images, for UAP generations. 
In addition, we also consider a data-limit setting, in which only 500 random images from the training set are sampled.
Aligning with previous work, we evaluate our method on the ImageNet validation set, which contains 50,000 images, using classical pre-trained CNN models AlexNet, GoogleNet, VGG16, VGG19, and ResNet152 as target models.

\noindent\textbf{Evaluation metrics:} We employ the widely used fooling ratio metric, which calculates the variation proportion of model predictions when applying the UAP, for evaluation.
% In the appendix,  we also include the results of an alternative evaluation metric, which measures the percentage of samples with prediction changes after applying UAP.

% \iffalse
\noindent\textbf{Comparative Methods:} 
% In the white-box attack scenario, we compare our method with existing approaches including UAP \cite{moosavi2017universal}, NAG \cite{mopuri2018nag}, GAP \cite{poursaeed2018generative}, DF-UAP \cite{zhang2020understanding}, Cos-UAP \cite{zhang2021data}, AT-UAP \cite{li2022learning}, and  SGA \cite{liu2023enhancing}. 
% For other settings, we compare our proposed method with UAP and GAP, as well as 
% SPGD, the  standard empirical risk maximization of \cref{eq:averaged-maximizing}.  
% Additionally, we evaluate two advanced methods: AT-UAP, which utilizes the optimal-data UAP maximin formulation of \cref{eq:adversarial-input-maximizing}, and SGA, which represents the current state-of-the-art method in this area.
% We regard SPGD as the baseline of AT-UAP, SGA, and the proposed DM-UAP since the latter three methods all build their  techniques upon the standard mini-batch training of the former.
% Their implementation details are included in Appendix-B.
% In the white-box attack scenario, we compare our method with the following existing approaches: UAP \cite{moosavi2017universal}, NAG \cite{mopuri2018nag}, GAP \cite{poursaeed2018generative}, DF-UAP \cite{zhang2020understanding}, Cos-UAP \cite{zhang2021data}, AT-UAP \cite{li2022learning}, and SGA \cite{liu2023enhancing}.
In the white-box attack scenario, we compare our method with the following existing approaches: UAP, NAG, GAP, DF-UAP, Cos-UAP, AT-UAP, and SGA.
For other settings, we compare DM-UAP with UAP, GAP, SPGD, AT-UAP, and SGA. SPGD is considered the baseline, as AT-UAP, SGA, and our proposed DM-UAP all build upon its standard mini-batch training.
% SPGD utilizes empirical risk maximization of \cref{eq:averaged-maximizing}.
% AT-UAP utilizes the optimal-data UAP maximin formulation of \cref{eq:adversarial-input-maximizing}.
Note that SGA is the current state-of-the-art method.

% \fi

\noindent\textbf{Hyper-parameters Setting:}  We set the maximum perturbation budget $\epsilon$ of all methods as $10/255$. Following SGA \cite{liu2023enhancing}, the number of training epochs $T$ is 20, and the batch size $B$ is 125.  The step numbers for inner model optimization $K_m$ and data optimization $K_d$ in our method are both 10, with default neighborhood size $\rho=1$ and $r=32$. 
% More details are provided in Appendix-A.

\begin{table}[!t]
\caption{The fooling ratio (\%) in the \textbf{white-box} setting by various UAP attack methods. The UAPs are crafted on the  AlexNet, GoogleNet, VGG16, VGG19, and ResNet152.}
\vspace{-2mm}
	\begin{center}

		\renewcommand{\arraystretch}{1}
		\scalebox{0.65}{
			\begin{tabular}{|c|cccccc|}
				\hline
				Method        & AlexNet        & GoogleNet      & VGG16          & VGG19          & ResNet152      & Average        \\ \hline \hline
				UAP           & 93.30           & 78.90           & 78.30           & 77.80           & 84.00           & 82.46          \\
			
				NAG           & 96.44          & 90.37          & 77.57          & 83.78          & 87.24          & 87.08          \\
				GAP           & -              & 82.70           & 83.70           & 80.10           & -              & 82.17          \\
				DF-UAP        & 96.17          & 88.94          & 94.30          & 94.98          & 90.08          & 92.89          \\
				Cos-UAP       & 96.50           & 90.50           & 97.40           & 96.40           & 90.20           & 94.20           \\
				AT-UAP        & 97.01          & 90.82          & 97.51          & 97.56          & 91.52          & 94.88          \\
				SGA & 96.99 & 90.64 & 97.83 & 96.56 & 92.86 & 94.98 \\ 
                \textbf{Ours} & \textbf{97.19} & \textbf{93.28} & \textbf{98.43} & \textbf{97.81} & \textbf{92.90} & \textbf{95.92} \\ 
                \hline
                
			\end{tabular}	
		}
	\end{center}
	
	\label{white-box-1w-untargeted}
 \vspace{-4mm}
\end{table}

\subsection{Generalization Performance of UAPs}
We  perform universal adversarial attacks under the white-box and black-box settings respectively and evaluate the overall performance of our proposed DM-UAP with baselines on the ImageNet validation set.

\textbf{White-box Attack.} We present the results of white-box attacks on five models using our DM-UAP approach, compared with other methods in \cref{white-box-1w-untargeted}. For SGA, we used its official source code and followed the settings from \cite{liu2023enhancing} with 10,000 samples. For other methods, we used results from their respective papers. Our method achieves the highest attack performance across all models. DM-UAP shows a notable improvement of over 2\% on GoogleNet. These indicate UAPs generated by our method can better generalize to unknown samples.

\textbf{Black-box Attack.} We evaluate transfer attacks with 10,000 and 500 training images. UAPs are generated for five considered models, and AEs are transferred between them. As shown in \cref{tab:transfer}, DM-UAP outperforms others across all models. Compared to AT-UAP and SGA, DM-UAP consistently achieves the highest average fooling ratio improvement over SPGD for all surrogate models, while AT-UAP and SGA are sometimes inferior to SPGD. With 10,000 images, DM-UAP's average fooling ratio improvement over SPGD ranges from 2.76\% to 11.15\%. With 500 images, DM-UAP's improvement ranges from 8.22\% to 45.19\%, compared to AT-UAP's -1.51\% to 38.00\% and SGA's 0.77\% to 22.55\%. DM-UAP outperforms SGA with an average increase in fooling ratio of 12.108\%.This manifests the importance of a dynamic model landscape proposed in our formulation in the limited samples scenario. 
% Visualization results are in Appendix B.

\begin{table}[!t]
    % \centering
   
    \caption{The fooling ratio (\%)  in the \textbf{ensemble-model} setting by different UAP generation methods. The UAPs are crafted on the ensemble models, \ie, AlexNet and VGG16.}
     \vspace{-5mm}
  
    \begin{center}
    \tabcolsep=0.1cm
      \renewcommand{\arraystretch}{1}
    \resizebox{0.48\textwidth}{!}{
    \begin{tabular}{|c|c|cc|ccccl|}
    \hline
         Method & Samples & AlexNet & VGG16 & GoogleNet & VGG19 &ResNet50 & ResNet152 &Average \\ \hline
          SPGD & \multirow{6}{*}{500} & 52.34* & 58.45* & 22.80 & 45.78  & 24.22 & 19.72 & 37.22 \\ 	 	 
         M-SPGD &  & 72.38* & 86.13* & 33.14 & 70.46 & 35.78 & 28.12 & 54.34$_{\textrm{\textcolor{green}{(+17.12)}}}$ \\
         AT-UAP & & 80.41*  & 92.11* & 42.28 & 80.29 & 43.59 & 33.49 & 62.03$_{\textrm{\textcolor{green}{(+24.81)}}}$ \\
         SGA &  & 82.34* & 90.70* & 51.07 & 78.79 & 50.95 & 40.76  & 65.77$_{\textrm{\textcolor{green}{(+28.55)}}}$ \\
         M-SGA &  & 85.93*  & 91.42* & 47.69 & 78.41 & 48.77 & 37.95  & 65.03$_{\textrm{\textcolor{green}{(+27.81)}}}$ \\ 	
         \textbf{Ours} &  & \textbf{91.39*} & \textbf{92.83*} & \textbf{56.23} & \textbf{82.74}  & \textbf{55.01} & \textbf{42.62}  & \textbf{70.14}$_{\textrm{\textcolor{green}{(+32.92)}}}$ \\ \cline{1-9} 	 
         SPGD & \multirow{6}{*}{10000} & 71.02* & 96.48* & 50.21 & 89.48 & 53.90 & 42.75 & 67.31 \\  
         M-SPGD &  & 73.35* & 97.47* & 52.18 & 91.42 & 56.19 & 44.07 & 69.11$_{\textrm{\textcolor{green}{(+1.80)}}}$ \\
         AT-UAP & &  82.33*  & 97.23* & 56.83 & 91.22 & 58.58 & 46.81 & 72.17$_{\textrm{\textcolor{green}{(+4.86)}}}$ \\
         SGA &  & 82.38* & 97.30* & 60.46 & 91.49 & 62.21 & 51.26 & 74.18$_{\textrm{\textcolor{green}{(+6.87)}}}$ \\
         M-SGA &  & 80.48* & \textbf{97.83*} & 60.23 & \textbf{92.46} & 61.86 & 50.51  & 73.89$_{\textrm{\textcolor{green}{(+6.58)}}}$ \\ 
         \textbf{Ours} &  & \textbf{91.35*} & 96.91*  & \textbf{66.34}  & 91.33 & \textbf{63.88} & \textbf{51.48} & \textbf{76.88}$_{\textrm{\textcolor{green}{(+9.57)}}}$\\ \hline
        
    \end{tabular}
    }
    \end{center}
    
    \label{tab:ensemble}
     \vspace{-5mm}
\end{table}

\vspace{-2mm}
\subsection{Scalability Performance of UAPs}
In this subsection, we analyze the scalability performance of the proposed method from various aspects, including  ensemble model setting,  diverse sample setting,  Transformer-to-CNN setting, and the attack-under-defense setting. 
% The first one is to evaluate whether our solution of dynamic maximin optimization can still operate in a more dynamic landscape, where multiple  DNNs are jointly used for the generation of UAPs.
% The other one is to evaluate whether the proposed method scales to different numbers of images used for UAP generations.

\begin{table*}[!t]
	\caption{The fooling ratio (\%) on five models in the \textbf{black-box} setting by different UAP attack methods. The UAPs are crafted on AlexNet, GoogleNet, VGG16, VGG19, and ResNet152, respectively. We conduct the evaluation using 10,000 and 500 images for training, respectively.
 The average improvement or deterioration of AT-UAP, SGA, and DM-UAP are also provided, with improvements highlighted in {\color{green} green} and deteriorations in  {\color{red} red}.
 * indicates the white-box model.}

      \vspace{-5mm}
      
	\begin{center}
		\renewcommand{\arraystretch}{1}
        \tabcolsep=0.06cm
		\resizebox{\textwidth}{!}{
			\begin{tabular}{|c|c|cccccl|cccccl|}
				\hline
				\multirow{2}{*}{Model}                                            & \multirow{2}{*}{Method} & \multicolumn{6}{|c|}{10,000 samples}  & \multicolumn{6}{|c|}{500 samples}\\ \cline{3-14} 
        & & AlexNet         & GoogleNet       & VGG16           & VGG19           & ResNet152       & Average   & AlexNet         & GoogleNet       & VGG16           & VGG19           & ResNet152       & Average     \\ \hline \hline
				\multirow{6}{*}{AlexNet}                         & UAP        & 84.28*          & 31.12           & 39.17           & 37.24           & 22.11           & 42.78     & 48.67* & 18.13 & 23.07 & 22.59 & 14.28 & 25.35      \\ 				
				& GAP       & 88.55*	& 35.52	& 52.85	& 49.22	& 28.81	& 50.99 & 83.96* & 32.99 & 48.26 & 45.17 & 26.98 & 47.47 \\ \cline{2-14}
				& SPGD   & 96.30*          & \textbf{54.16}           & 61.39           & 58.89           & 36.78           & 61.50        & 89.20* & 40.79 & 51.15 & 49.53 & 28.05 & 51.74  \\ 
				& AT-UAP   & 96.74* & 48.86 & 62.23 & 58.80 & 33.36 & 60.00$_{\textrm{\textcolor{red}{(-1.50)}}}$    & 93.80* & 31.78 & 51.98 & 48.57 & 25.00 & 50.23$_{\textrm{\textcolor{red}{(-1.51)}}}$ \\
				& SGA     & 96.99* & 46.62  & 65.57  & 59.81  & 34.30  & 60.66$_{\textrm{\textcolor{red}{(-0.84)}}}$  & 94.91* & 34.88 & 55.27 & 50.85 & 26.64 & 52.51$_{\textrm{\textcolor{green}{(+0.77)}}}$ \\ 
                & \textbf{Ours}    & \textbf{97.19*} & 53.95  & \textbf{68.20}  & \textbf{63.03}  & \textbf{38.91}  & \textbf{64.26}$_{\textrm{\textcolor{green}{(+2.76)}}}$  & \textbf{96.13*} & \textbf{48.11} & \textbf{63.28} & \textbf{58.38} & \textbf{33.92} & \textbf{59.96}$_{\textrm{\textcolor{green}{(+8.22)}}}$ \\ \hline
                																									
				\multirow{6}{*}{GoogleNet}                       & UAP        & 39.55	& 55.01*	& 49.82	& 49.11	& 29.66	& 44.63         & 23.83 & 15.91* & 19.45 & 18.94 & 11.78 & 17.98  \\
				& GAP        & 53.31	& 80.21*	& 72.56	& 70.62	& 49.85	& 65.31        & 37.59 & 33.32* & 35.63 & 35.41 & 20.81 & 32.55   \\ \cline{2-14}
				& SPGD   & 50.47           & 86.09* & 66.79           & 65.93           & 44.66           & 62.79         & 31.28 & 55.35* & 29.50 & 28.69 & 17.89 & 32.54 \\
                    & AT-UAP   & 54.13 & 92.55* & 79.13 & 76.54 & 53.51 & 71.17$_{\textrm{\textcolor{green}{(+8.38)}}}$              & 49.49 & 80.63* & 65.03 & 63.23 & 41.14 & 59.90$_{\textrm{\textcolor{green}{(+27.36)}}}$\\
				& SGA    & \textbf{59.22}  & 88.46*          & 77.10  & 74.80  & 54.97  & 70.91$_{\textrm{\textcolor{green}{(+8.12)}}}$ & 50.77 & 68.10* & 60.18 & 59.13 & 37.28 & 55.09$_{\textrm{\textcolor{green}{(+22.55)}}}$ \\ 
                 & \textbf{Ours}    & 57.62 & \textbf{93.28*}  & \textbf{81.45}  & \textbf{80.05}  & \textbf{57.29}  & \textbf{73.94}$_{\textrm{\textcolor{green}{(+11.15)}}}$ & \textbf{57.29} & \textbf{88.07*} & \textbf{76.88} & \textbf{74.57} & \textbf{51.43} & \textbf{69.65}$_{\textrm{\textcolor{green}{(+37.11)}}}$ \\ \hline
                										 															
				\multirow{6}{*}{VGG16}                           & UAP        & 33.33	& 36.19	& 75.36*	& 64.09	& 31.33	& 48.06          & 22.87 & 14.97 & 26.67* & 22.84 & 12.72 & 20.01  \\
				& GAP        & 35.90	& 50.05	& 84.59*	& 76.49	& 38.97	& 57.20      & 34.65 & 26.39 & 57.82* & 44.69 & 21.39 & 37.00       \\ \cline{2-14}
				& SPGD   & 40.67           & 43.46           & 92.81*          & 83.18           & 44.44           & 60.91    & 34.57 & 24.10 & 73.10* & 56.04 & 21.06 & 41.77       \\
                    & AT-UAP   & 43.72 & 42.39 & 96.68* & 88.13 & 34.97 & 61.18$_{\textrm{\textcolor{green}{(+0.27)}}}$                 & 43.34 & 33.48 & 90.13* & 75.06 & 28.84 & 54.17$_{\textrm{\textcolor{green}{(+12.40)}}}$   \\
				& SGA    & 44.48  & 52.53  & 97.83* & 92.36  & 48.30  & 67.10$_{\textrm{\textcolor{green}{(+6.19)}}}$ & 42.85 & 41.96 & 92.64* & 80.83 & 36.24 & 58.90$_{\textrm{\textcolor{green}{(+17.13)}}}$ \\ 
                & \textbf{Ours}    & \textbf{49.35} & \textbf{57.18}  & \textbf{98.43*}  & \textbf{94.12}  & \textbf{49.97}  & \textbf{69.81}$_{\textrm{\textcolor{green}{(+8.90)}}}$ & \textbf{45.99} & \textbf{49.96} & \textbf{96.78*} & \textbf{89.03} & \textbf{43.57} & \textbf{65.07}$_{\textrm{\textcolor{green}{(+23.30)}}}$ \\ \hline
 																									
				\multirow{6}{*}{VGG19}                           & UAP        & 33.98	& 36.62	& 64.57	& 74.77*	& 30.48	& 48.08        & 23.59 & 15.26 & 25.13 & 26.09* & 12.77 & 20.57     \\
				& GAP        & 45.23	& 43.69	& 73.89	& 82.31*	& 30.02	& 55.03     & 38.86 & 30.36 & 52.12 & 60.83* & 23.19 & 41.07      \\ \cline{2-14}
				& SPGD   & 40.76     & 47.84	& 84.26	& 92.90*	& 45.43	& 62.24      & 33.38 & 22.30 & 52.50 & 65.49* & 19.79 & 38.69        \\
                & AT-UAP   & 45.42 & 43.72 & 90.61 & 95.47* & 40.05 & 63.05$_{\textrm{\textcolor{green}{(+0.81)}}}$            & 42.82 & 34.20 & 77.99 & 88.02* & 29.89 & 54.58$_{\textrm{\textcolor{green}{(+15.89)}}}$         \\
				& SGA    & 45.89  & 55.53  & 92.62  & 96.56* & 49.90  & 68.10$_{\textrm{\textcolor{green}{(+5.86)}}}$ & 42.70 & 42.90 & 84.42 & 90.88* & 36.69 & 59.52$_{\textrm{\textcolor{green}{(+20.83)}}}$ \\ 
                & \textbf{Ours}    & \textbf{49.81} & \textbf{59.91}  & \textbf{95.51}  & \textbf{97.81*}  & \textbf{55.75}  & \textbf{71.76}$_{\textrm{\textcolor{green}{(+9.52)}}}$ & \textbf{47.24} & \textbf{50.97} & \textbf{91.33} & \textbf{96.55*} & \textbf{41.57} & \textbf{65.53}$_{\textrm{\textcolor{green}{(+26.84)}}}$ \\ \hline
 																					
				\multicolumn{1}{|l|}{\multirow{6}{*}{ResNet152}} & UAP        & 35.15	& 36.04	& 49.19	& 47.50	& 57.36*	& 45.05     & 27.28 & 18.06 & 25.02 & 24.10 & 17.97* & 22.49        \\
				\multicolumn{1}{|l|}{}                           & GAP        & 47.70	& 56.08	& 68.63	& 66.51	& 73.80*	& 62.54     & 39.60 & 38.23 & 46.90 & 46.24 & 46.66* & 43.53        \\  \cline{2-14}
				\multicolumn{1}{|l|}{}                           & SPGD   & 46.12	& 55.01	& 77.33	& 74.19	& 90.33*	& 68.60      & 30.23 & 19.01 & 27.33 & 26.40 & 18.87* & 24.37     \\
                    \multicolumn{1}{|l|}{}                           & AT-UAP   & 47.69 & 57.94 & 77.88 & 75.09 & 92.07* & 70.13$_{\textrm{\textcolor{green}{(+1.53)}}}$           & 44.67 & 47.62 & 71.56 & 67.51 & 80.49* & 62.37$_{\textrm{\textcolor{green}{(+38.00)}}}$ \\
				\multicolumn{1}{|l|}{}                           & SGA    & 49.30  & 62.40  & 80.53  & 78.18  & 92.86* & 72.65$_{\textrm{\textcolor{green}{(+4.05)}}}$ & 40.53 & 35.57 & 49.45 & 47.51 & 43.01* & 43.21$_{\textrm{\textcolor{green}{(+18.84)}}}$ \\ 
                 \multicolumn{1}{|l|}{}                           & \textbf{Ours}    & \textbf{50.09}  & \textbf{63.50}  & \textbf{81.49}  & \textbf{79.10}  & \textbf{92.90*} & \textbf{73.42}$_{\textrm{\textcolor{green}{(+4.82)}}}$  & \textbf{51.26} & \textbf{59.17} & \textbf{78.05} & \textbf{75.40} & \textbf{83.92*} & \textbf{69.56}$_{\textrm{\textcolor{green}{(+45.19)}}}$  \\ \hline

			\end{tabular}
		}
	\end{center}

	\label{tab:transfer}
\vspace{-3mm}
\end{table*}

\textbf{Ensemble-Model Setting}. Following \cite{liu2023enhancing}, we implement the model ensemble method using the  averaged loss functions of two models, \ie, AlexNet and VGG16. 
In this experiment, we still use SPGD as the baseline method,  and compare the improvement of DM-UAP  with those of others. 
For more comparison, we also integrate   SPGD and SGA with the momentum \cite{dong2018boosting} method,  denoted as M-SPGD and M-SGA.
In addition to the commonly used five models, we also test the transferability on ResNet50. 
The results are reported in \cref{tab:ensemble}.  
DM-UAP still outstrips the others by a clear superiority, which verifies the effectiveness of our dynamic maximin formulation still suffices in such a more complex optimization landscape.

\begin{figure}[h]
  \centering
\vspace{-2mm}
  \includegraphics[width=0.35\textwidth]{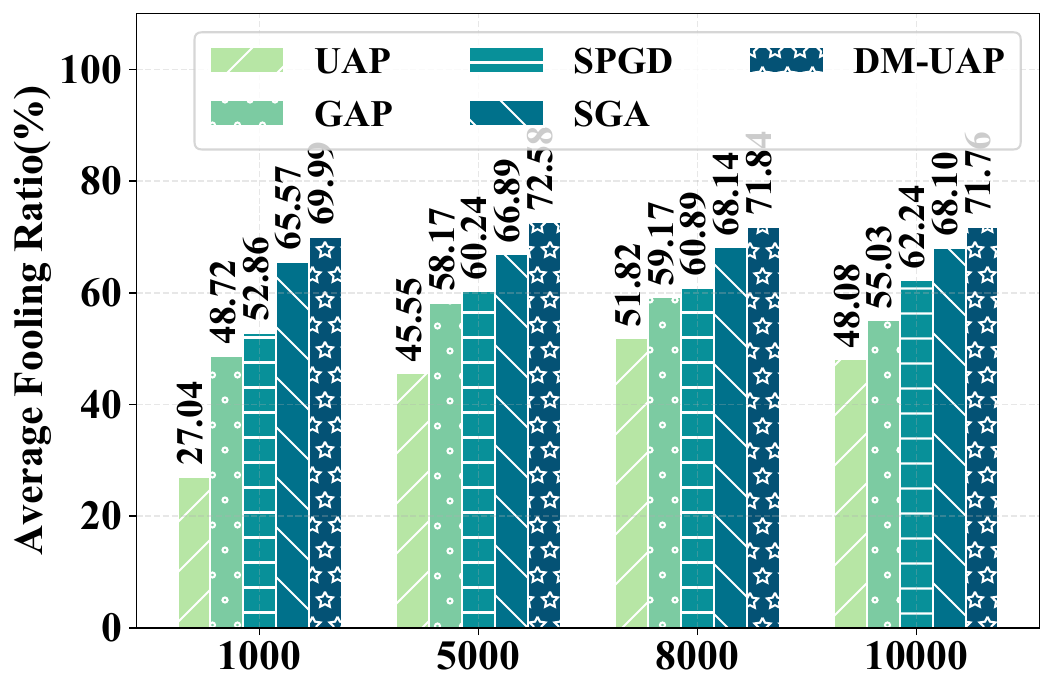}
  \caption{Average fooling ratio (\%) on five models in the \textbf{diverse-sample} training scenarios. The UAPs are crafted by UAP, GAP, SPGD, and SGA, and our DM-UAP on VGG19.}
    \label{fig:diverse}
    \vspace{-5mm}
\end{figure}

\textbf{Diverse-Sample Setting}. Furthermore, we investigate the impact of varying the number of training samples on the attack performance. The results are depicted in  \cref{fig:diverse}. 
Our method consistently achieves the best performance with any number of training samples. However, we observe that once the number of samples exceeds a certain threshold, the corresponding increase in attack performance plateaus.
The phenomenon underscores the limitations of excessive increase in the number of training samples.
% , since as the number of training samples increases, the optimization task becomes  more challenging.

\textbf{Transformer-to-CNN Setting.} 
Unlike CNN models, transformer models employ self-attention mechanisms rather than convolutional blocks. Studies indicate that transformer models demonstrate reduced cross-model transferability to other architectures, particularly CNN. We adapt transformer models from DeiT \cite{touvron2021training} and ViT \cite{dosovitskiy2020image} family as surrogate models and evaluate the cross-model performance of different UAP methods. 
We can see from \cref{tab:vit}, that compared to SPGD, AT-UAP and SGA, our method consistently achieves the highest fooling ratio against all black-box CNN models.

\begin{table}[!t]
    % \centering
   
    \caption{Fooling ratio (\%) in the \textbf{Transformer-to-CNN} setting
    using different generation methods.}
     \vspace{-2mm}
  
    \begin{center}
    \tabcolsep=0.09cm
      \renewcommand{\arraystretch}{1}
    \resizebox{0.48\textwidth}{!}{
    \begin{tabular}{|c|c|cccccl|}
    \hline
        Model & UAP Method & AlexNet & VGG16 & GoogleNet & VGG19  & ResNet152 &Average \\ \hline
        \multirow{4}{*}{ViT-B} & SPGD  & 51.65 & 30.10 & 49.39 & 47.66  & 23.86  & 40.53 \\ 	 	  
        & SGA   & 54.31 & 37.86 & 54.18 & 52.26 & 28.93 & 45.51$_{\textrm{\textcolor{green}{(+4.98)}}}$ \\
        & AT-UAP & 56.24 & 44.88 & 55.47 & 54.19 & 32.31 & 48.62$_{\textrm{\textcolor{green}{(+8.09)}}}$	\\
        & \textbf{Ours}   & \textbf{59.01} & \textbf{51.63} & \textbf{62.35} & \textbf{61.62}  & \textbf{38.31}  & \textbf{54.58}$_{\textrm{\textcolor{green}{(+14.05)}}}$ \\ \cline{1-8} 	 
        \multirow{4}{*}{ViT-L} & SPGD  & 35.94 & 24.93 & 36.01 & 34.83  & 18.55 & 30.05 \\  
        
        & SGA   & 43.34 & 34.69 & 47.18 & 46.35 & 24.60  & 39.23$_{\textrm{\textcolor{green}{(+9.18)}}}$ \\
        & AT-UAP & 51.38 & 40.14 & 53.73 & 51.47 & 28.63 & 45.07$_{\textrm{\textcolor{green}{(+15.02)}}}$  \\ 
        & \textbf{Ours}  & \textbf{54.94} & \textbf{46.55}    & \textbf{59.50} & \textbf{56.26} & \textbf{32.52} & \textbf{49.95}$_{\textrm{\textcolor{green}{(+19.90)}}}$\\

        \cline{1-8} 	 
        \multirow{4}{*}{DeiT-S} & SPGD & 34.38 & 21.76  & 41.09 & 37.47 & 19.53 & 30.85 \\  
        
        & SGA   & 48.35 & 40.69 & 49.63 & 47.43  & 28.37 & 42.89$_{\textrm{\textcolor{green}{(+12.04)}}}$ \\
        & AT-UAP & 54.29 & 46.59 & 57.36 & 55.11 & 32.21 & 49.11$_{\textrm{\textcolor{green}{(+18.26)}}}$ \\
        & \textbf{Ours}   & \textbf{56.60}   & \textbf{50.09}  & \textbf{62.00} & \textbf{59.64} & \textbf{37.35} & \textbf{53.14}$_{\textrm{\textcolor{green}{(+22.29)}}}$\\
        \cline{1-8} 
         \multirow{4}{*}{DeiT-B} & SPGD & 39.66 & 26.72  & 40.02 & 38.04 & 19.66 & 32.82 \\  
        
        & SGA   & 43.30 & 27.81 & 44.65 & 42.52  & 21.18 & 35.89$_{\textrm{\textcolor{green}{(+3.07)}}}$ \\
        & AT-UAP & 41.12 & 27.37 & 48.41 & 44.44 & 21.74 & 36.62$_{\textrm{\textcolor{green}{(+3.80)}}}$  \\
        & \textbf{Ours}   & \textbf{46.32}   & \textbf{29.56}  & \textbf{51.26} & \textbf{47.12} & \textbf{23.22} & \textbf{39.50}$_{\textrm{\textcolor{green}{(+6.68)}}}$\\
        \hline

    \end{tabular}
    }
    \end{center}
    
    \label{tab:vit}
     \vspace{-2mm}
\end{table}

\begin{table}[!t]
    % \centering
    \caption{The fooling ratio (\%) in the \textbf{attack-under-defense} setting on five models by different UAP generation methods.}
     \vspace{-5mm}
  
    \begin{center}
    \tabcolsep=0.04cm
      \renewcommand{\arraystretch}{1}
    \resizebox{0.48\textwidth}{!}{
    \begin{tabular}{|c|c|cccccl|}
    \hline
        Defense & UAP Method  & AlexNet & VGG16 & GoogleNet & VGG19  & ResNet152 &Average \\ \hline
        \multirow{3}{*}{JPEG} & SPGD  & 91.9 & 79.6 & 64.6 & 60.0  & 42.6  & 67.7 \\ 	 	 
        
        & SGA &   91.3 & \textbf{82.2} & 81.5 & 81.3 & \textbf{56.6} & 78.6$_{\textrm{\textcolor{green}{(+10.9)}}}$ \\
        	
        & \textbf{Ours} &   \textbf{93.0} & 79.5 & \textbf{89.6} & \textbf{84.6}  & 53.7  & \textbf{80.1}$_{\textrm{\textcolor{green}{(+12.36)}}}$ \\ \cline{1-8} 	 
        \multirow{3}{*}{NRP} & SPGD  & 61.2 & 40.7 & 56.3 & 52.9 & 35.8 & 49.4 \\  
        
        & SGA &   61.8 & 40.1 & 55.1 & 55.1 & 36.5  & 49.7$_{\textrm{\textcolor{green}{(+0.3)}}}$ \\
         
        & \textbf{Ours} &  \textbf{63.1} & \textbf{41.3}    & \textbf{56.1} & \textbf{55.3} & \textbf{38.6} & \textbf{50.9}$_{\textrm{\textcolor{green}{(+1.5)}}}$\\

        \cline{1-8} 	 
        \multirow{3}{*}{Diffpure} & SPGD  & 36.0 & 38.2  & 36.7 & 38.2 & 39.7 & 37.8 \\  
        
        & SGA   & \textbf{39.4} & 37.5 & 37.9 & 36.8  & 37.4 & 37.8$_{\textrm{\textcolor{green}{(+0)}}}$ \\
         
        & \textbf{Ours}  & 38.5   & \textbf{38.8}  & \textbf{39.3} & \textbf{40.8} & \textbf{41.0} & \textbf{39.7}$_{\textrm{\textcolor{green}{(+1.9)}}}$\\
        \hline
        
    \end{tabular}
    }
    \end{center}
     \vspace{-6mm}
     \label{tab:defense}
\end{table}

\begin{figure*}[!t]
  \centering
  \begin{subfigure}[b]{0.23\textwidth}
    \includegraphics[width=\linewidth]{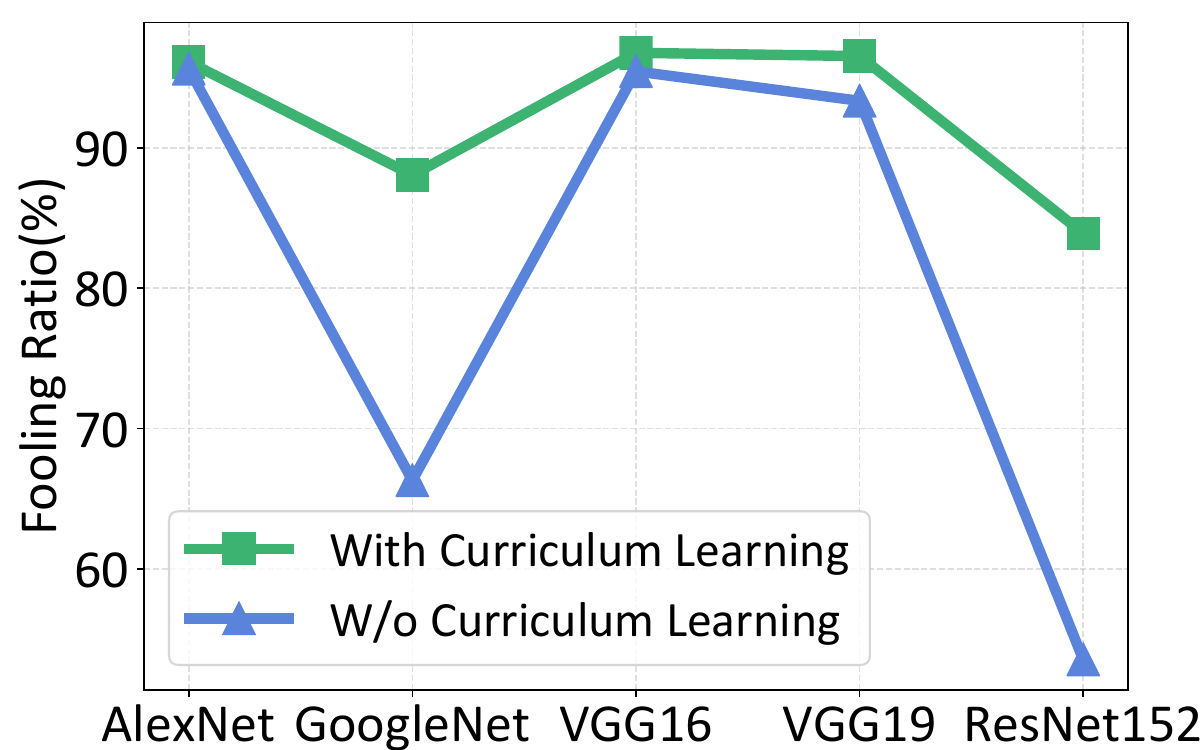} 
    \caption{Curriculum Learning}
    \label{fig:curriculum}
  \end{subfigure}
    \hfill 
  \begin{subfigure}[b]{0.23\textwidth}
    \includegraphics[width=\linewidth]{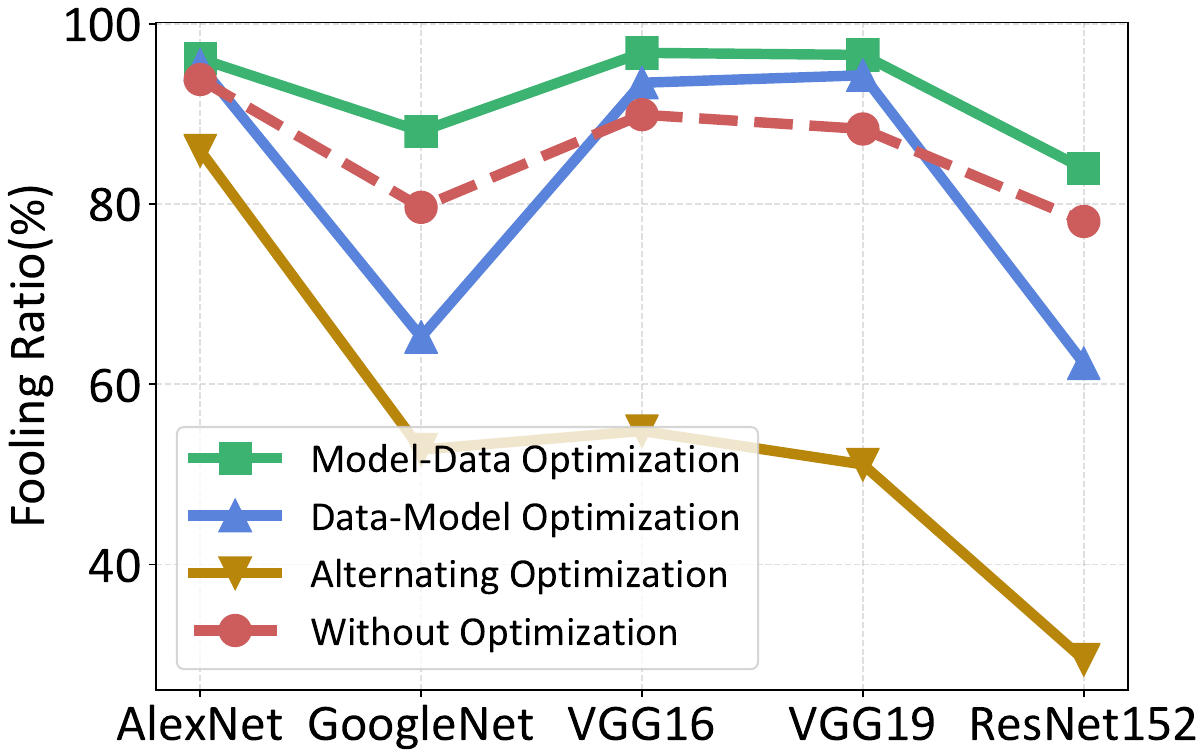} 
    \caption{Optimization order}
    \label{fig:order_ablation}
  \end{subfigure}
    \hfill 
  \begin{subfigure}[b]{0.23\textwidth}
    \includegraphics[width=\linewidth]{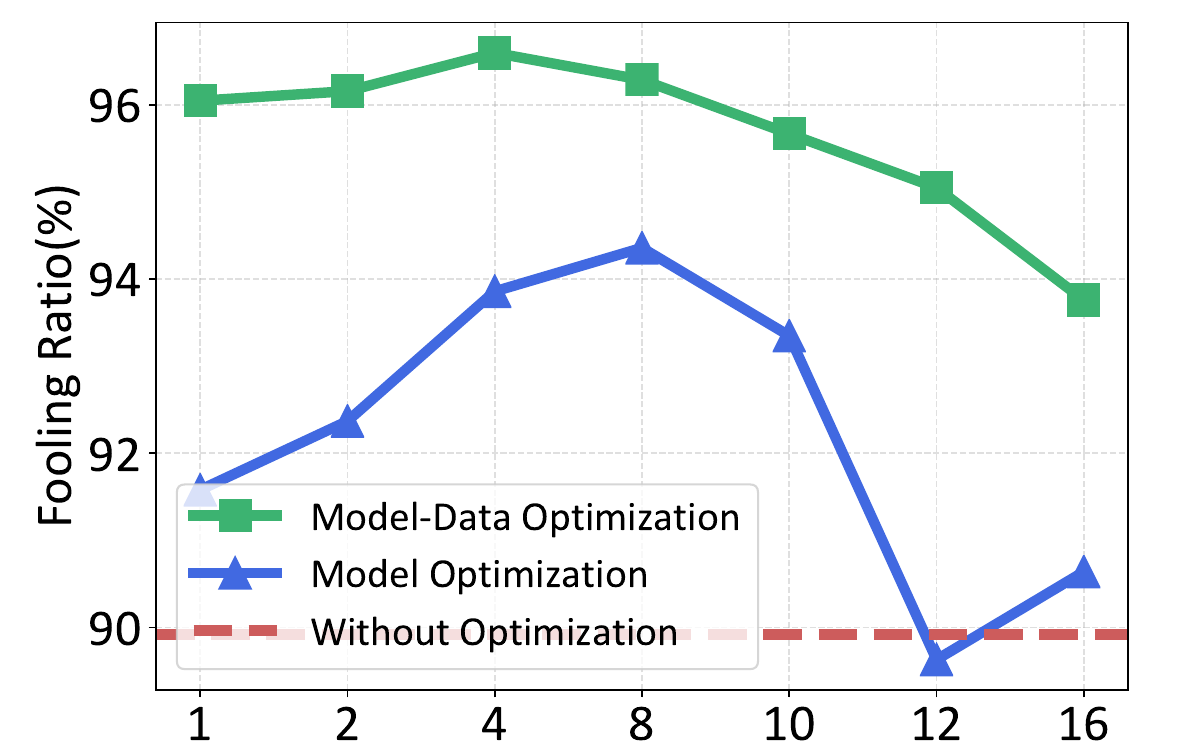}    
    \caption{Model neighborhood size}
    \label{fig:theta_ablation}
  \end{subfigure}
  \hfill 
  \begin{subfigure}[b]{0.23\textwidth}
    \includegraphics[width=\linewidth]{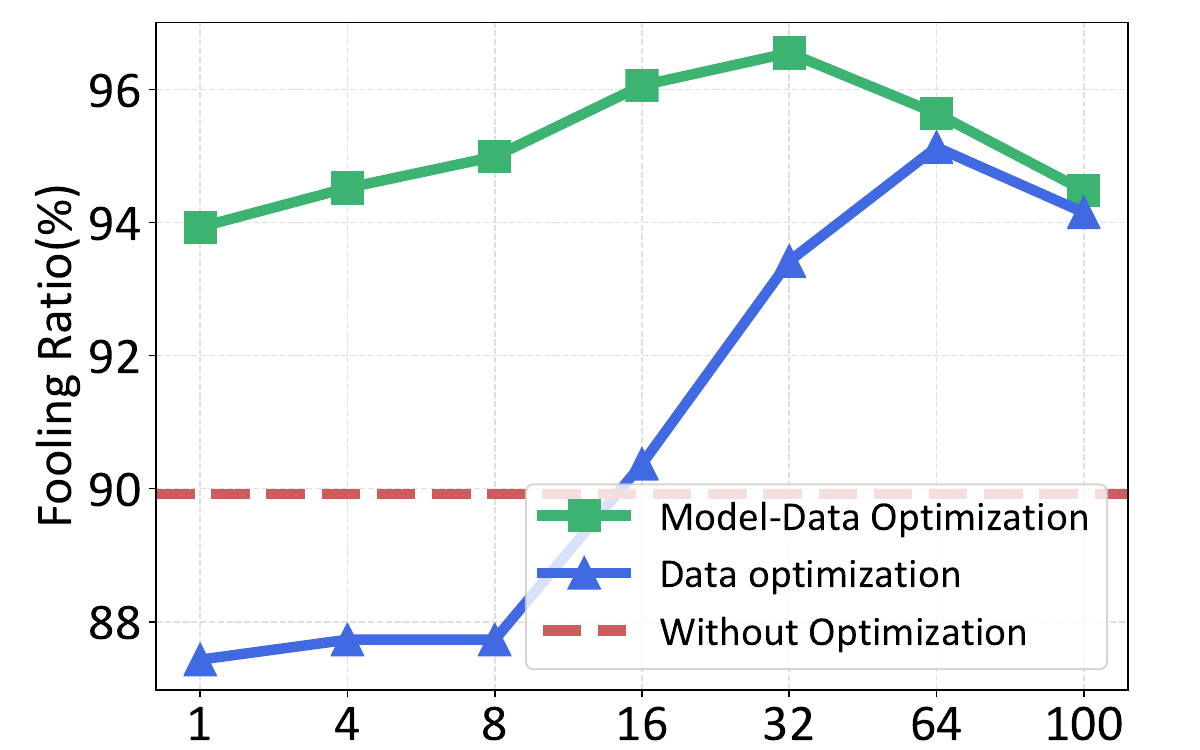}
    \caption{Data neighborhood size}
    \label{fig:x_ablation}
  \end{subfigure}
  
  \vspace{-2mm}
  
  \caption{
Ablation study on model and data optimization. (a) Fooling ratios of DM-UAP with/without curriculum learning, \textit{i.e}, increasing neighborhood sizes. (b) Fooling ratio by different optimization orders in white-box setting for five models. (c) Fooling ratios of DM-UAP with/without data optimization for different model neighborhood sizes. (d) Fooling ratios of DM-UAP with/without model optimization for different data neighborhood sizes.
}
  \label{ablation}
  \vspace{-5mm}
\end{figure*}

\textbf{Attack-under-Defense Setting.}
We evaluate the attack performance of AEs crafted by different UAP methods under three common defenses: JPEG \cite{das2018shield}, NPR \cite{Naseer_2020_CVPR}, and DiffPure\cite{nie2022diffusion}. JPEG uses lossy compression to remove adversarial perturbations. NPR trains a purifier network that minimizes the difference in perceptual features between clean and adversarial images. DiffPure uses Gaussian noise to smooth out adversarial perturbations and then denoises images using a pre-trained diffusion model. \cref{tab:defense} shows that DM-UAP performs the best on average under every defense.

\subsection{Ablation Study}
We conduct ablation studies to evaluate the impact of curriculum learning, the model and data optimization order, and optimization neighborhood sizes on the performance of our proposed framework, using 500 training images by default.

\textbf{On the curriculum UAP learning.} 
We examine the impact of curriculum learning by gradually increasing the model neighborhood size $\rho$ from 0 to 4 and the data neighborhood size $r$ from 0 to 32 throughout the training epochs. 
 For comparison, fixed neighborhood sizes of $\rho = 4$ and $r = 32$ are also used. 
 As shown in \cref{fig:curriculum}, the results underscore the importance of using an easy-to-hard antagonistic approach as in curriculum learning for optimizing UAP, which significantly enhances its effectiveness. 

\textbf{On the optimization order of model and data.} 
We explore the influence of different optimization orders to highlight the  ``optimize model first, data second'' sequencing adopted in DM-UAP.
In \cref{fig:order_ablation},   we use ``Model-Data Optimization" to denote the process of  10 model optimization steps followed by 10 data optimization steps as in DM-UAP. 
 In contrast, ``Data-Model Optimization" reverses this order. ``Alternating Optimization" means alternating single optimization steps between the model and data 20 times.
``Without Optimization" means both model and data are not optimized, which reduces to the formulation of \cref{eq:averaged-maximizing}. 
The results show ``Model-Data Optimization" consistently outperforms the ``Without Optimization" baseline. 
``Data-Model Optimization" sometimes, while ``Alternating Optimization" consistently underperforms.
These confirm our hypothesis that the model should be optimized using unperturbed data.

\textbf{On the model neighborhood size.}  We explore the influence of the maximum model neighborhood size, with data neighborhood size set as $r = 32$ for ``Model-Data Optimization". We tested $\rho$ values of 1, 2, 4, 8, 10, 12, and 16, as depicted in \cref{fig:theta_ablation}. 
For comparison, the formulation of optimal-parameters UAP maximin in \cref{eq:adversarial-parameter-maximizing}  is also assessed, denoted as ``Model Optimization''.
The results indicate that ``Model-Data Optimization" consistently surpasses ``Model Optimization", and  their optimal $\rho$ value differ.

\textbf{On the data neighborhood size.}  Similarly, we explore the influence of the maximum data neighborhood size $r$, with the model neighborhood size set as $\rho = 4$ for ``Model-Data Optimization". 
We test  $r$ values of 1, 4, 8, 16, 32, 64, and 100, illustrated in \cref{fig:x_ablation}.
``Data Optimization", the formulation of optimal-data  UAP maximin in \cref{eq:adversarial-input-maximizing} is also examined by varying $r$.
The results also demonstrate that ``Model-Data Optimization" invariably outperforms ``Data Optimization", their  optimal $r$ value differ.

\section{Limitation and Future Work}
% Our DM-UAP introduces a novel maximin formulation to improve the generalization of UAP, which involves the dynamic optimization of both model and data in the UAP generation process. 
% As a result, this will lead to extra computational expenses compared to existing methods.
% Specifically, the time expense of DM-UAP is roughly 1.6 times of  SGA and twice that of AT-UAP (see Appendix-C). Nevertheless,  considering the characteristics of universal adversarial perturbation, the computational costs can be  negligible. 
% Once the UAPs are crafted, there is no need to spend additional time and resources generating corresponding perturbations for each sample. The Off the shelf UAPs can be used to produce AEs at scale.

Our DM-UAP introduces a novel maximin formulation to enhance the generalization of UAP by dynamically optimizing both the model and data during the UAP generation process. 
As a result, this incurs additional computational expenses. Specifically, for crafting UAPs on VGG16, the time expense for DM-UAP is approximately 1.6 times that of SGA and twice that of AT-UAP (see \cref{tab:resource consumption}). 
However, considering the universal nature of UAP, these computational costs can be deemed negligible. 
Once the UAPs are created, there is no need for additional computations, as the off-the-shelf UAPs can be readily utilized to generate AEs at scale. 
For future studies on crafting UAPs against large models that require more memory, we think adaptively selecting part of the parameters for optimization may be worth exploring.

\begin{table}[h]
    \centering
    \caption{The optimization expenses of different UAP methods. UAPs are obtained from the VGG16 model.}
        \vspace{-3mm}
    \resizebox{0.45 \textwidth}{!}{%
    \begin{tabular}{|c|c|c|c|}
        \hline
       UAP method  & Time consumption & Memory consumption & \makecell[c]{Average attack \\ success rate(\%)}  \\ \hline
       SPGD  & 27min'58s & 19620 MiB &  60.91\\	\hline
       AT-UAP  & 	1h'12min'34s & 20210 MiB & 61.18\\ \hline
       SGA  & 	1h'28min'44s & 10382 MiB & 67.10\\ \hline
       DM-UAP(Ours)  & 	2h'21min'26s & 20714 MiB & 69.81\\ \hline
    \end{tabular}
    }

    \label{tab:resource consumption}
    \vspace{-3mm}
\end{table}

\section{Conclusion}

% In this paper, we introduce a novel formulation that generates UAP  through a  dynamic maximin optimization strat
% Compared to previous methods, our  DM-UAP method  innovatively leverages dynamic  model parameters during the optimization process of UAPs, marking the first exploration into  the model parameter landscape for  UAP generation.
% To address the dynamic maximin formulation effectively, we proposed an iterative max-min-min optimization framework that optimizes the model-data pairs to minimize classification loss  on the fly within each mini-batch iteration, as well as a curriculum learning algorithm designed to fully explore the joint landscape of model parameters and data. 
% Experimental results on the ImageNet dataset under various settings validate the proposed DM-UAP can significantly improve the generalization of UAPs  and exceed the state-of-the-art methods by  clear superiority.

In this paper, we present DM-UAP, a novel approach for generating UAPs with a dynamic maximin optimization strategy.
DM-UAP not only optimizes the data used for training but also takes into account dynamic model parameters.
Specifically, DM-UAP incorporates an iterative max-min-min optimization framework that dynamically minimizes classification loss to obtain model-data pairs, as well as a curriculum learning algorithm to thoroughly explore the combined landscape of model parameters and data.
Extensive experiments on ImageNet demonstrate the superior performance of DM-UAP over state-of-the-art methods, significantly improving the generalization of generated UAPs.
% We introduce DM-UAP, a new method for generating UAPs with a dynamic maximin optimization strategy. It optimizes training data and dynamic model parameters. Our approach includes an iterative max-min-min framework and a curriculum learning algorithm for thorough exploration of the model-data landscape. Experiments on ImageNet show its superior performance and improved generalization over existing methods.

\section*{Acknowledgement}
Minghui's work is supported by the National Natural Science
Foundation of China (Grant No.62202186). 
Shengshan's work is supported by the National Natural Science Foundation of China (Grant No.62372196). Minghui Li is the corresponding author.

\bibliography{aaai25}

\end{document}